\documentclass[10pt,twocolumn,letterpaper]{article}

\usepackage{iccv}
\usepackage{times}
\usepackage{epsfig}
\usepackage{graphicx}
\usepackage{amsmath}
\usepackage{amssymb}

\usepackage{booktabs}
\usepackage{pifont}
\usepackage{floatrow}
\usepackage{caption}
\usepackage{subcaption}
\usepackage{multicol}

\newfloatcommand{capbtabbox}{table}[][\FBwidth]

\usepackage[pagebackref=true,breaklinks=true,letterpaper=true,colorlinks,bookmarks=false]{hyperref}

\iccvfinalcopy 

\def\iccvPaperID{7} 

\ificcvfinal\fi
\begin{document}

\title{Hyperparameter-Free Losses for Model-Based Monocular Reconstruction}

\author{Eduard Ramon\\
Crisalix SA\\
{\tt\small eduard.ramon@crisalix.com}
\and
Guillermo Ruiz\\
Crisalix SA\\
{\tt\small guillermo.ruiz@crisalix.com}
\and
Thomas Batard\\
Crisalix SA\\
{\tt\small thomas.batard@crisalix.com}
\and
Xavier Gir\'{o}-i-Nieto\\
Universitat Polit\`{e}cnica de Catalunya\\
{\tt\small xavier.giro@upc.edu}
}

\maketitle

\begin{abstract}

This work proposes novel hyperparameter-free losses for single view 3D reconstruction with morphable models (3DMM). We dispense with the hyperparameters used in other works by exploiting geometry, so that the shape of the object and the camera pose are jointly optimized in a sole term expression. This simplification reduces the optimization time and its complexity. Moreover, we propose a novel implicit regularization technique based on random virtual projections that does not require additional 2D or 3D annotations. Our experiments suggest that minimizing a shape reprojection error together with the proposed implicit regularization is especially suitable for applications that require precise alignment between geometry and image spaces, such as augmented reality. We evaluate our losses on a large scale dataset with 3D ground truth and publish our implementations to facilitate reproducibility and public benchmarking in this field.

\end{abstract}

\section{Introduction}

Inferring the geometry of objects from a single or multiple images is a well-studied problem by the computer vision community. Traditionally, the employed techniques have been based in geometry and/or photometry \cite{horn1970shape, westoby2012structure}, which usually require a large amount of images in order to create precise reconstructions.
Recently, the capacity of deep neural networks \cite{goodfellow2016deep} to obtain hierarchical representations of the images and to encode prior knowledge has been applied to 3D reconstruction in order to learn the implicit mapping between images and geometry \cite{choy20163d, yan2016perspective}.

\begin{figure}
    \centering
    \includegraphics[width=8cm]{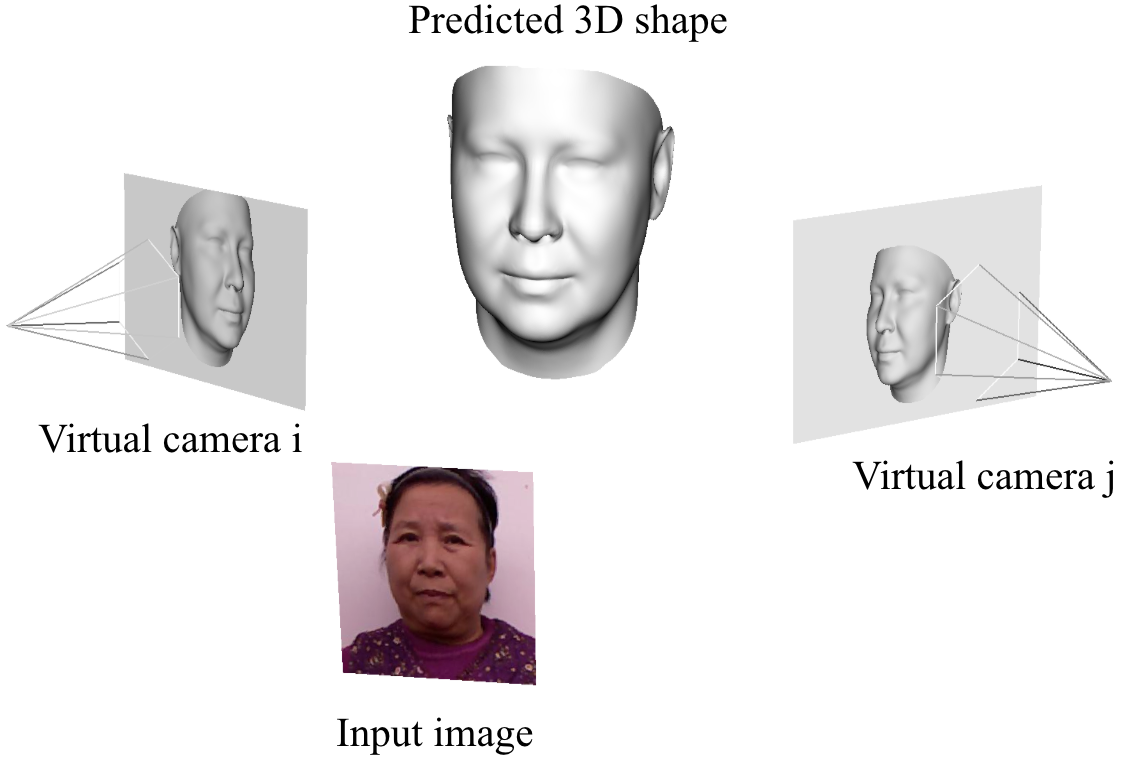}
    \caption{Overview of our random projections approach for implicit 3D shape regularization.}
    \label{fig:overview}
\end{figure}

Nevertheless, employing deep neural networks to solve 3D related problems implies some specific issues that need to be addressed. One of the main drawbacks is the 3D data representation. The trivial generalization from 2D images to 3D space are the 3D voxel grids. This representation, which is simple and allows the use of 3D convolutions, does an inefficient use of the target space when trying to reconstruct surfaces. Moreover, state of the art methods that use this representation mostly work at resolutions around 128x128x128 voxels \cite{choy20163d, yan2016perspective}, which are too small for most of the applications. 3D meshes \cite{kanazawa2018learning,wang2018pixel2mesh} are a more convenient representation because they efficiently model surfaces and can be easily textured and animated for computer graphics applications. However, 3D meshes are defined in a non-Euclidean space, where the usual deep learning operations like convolutions are not defined. Geometric deep learning \cite{bronstein2017geometric}  is nowadays a hot research area to bring basic operations to non-Euclidean domains like graphs and manifolds, which is the case of 3D meshes. Finally, 3D Morphable Models (3DMM) \cite{blanz1999morphable} are used for category-specific problems to reduce the dimensionality of plausible solutions and lead to more robust and likely predictions.

Another challenge when working on 3D reconstruction using deep learning is the lack of labelled data. In tasks like image recognition, there exist large annotated datasets with millions of images \cite{deng2009imagenet}. Unfortunately, the data is not as abundant in 3D as it is in 2D and, consequently, researchers have walked around this limitation with different strategies. Defining losses in the image domain \cite{tewari2017mofa,richardson2017learning} is a common approach since it provides flexibility to use different kinds of 2D annotations like sparse sets of keypoints, foreground masks or pixel intensities. A second strategy is the use of synthetic data \cite{richardson20163d,richardson2017learning,roth2014unconstrained} since it provides perfect 3D ground truth. Unfortunately, those systems trained with synthetic data tend to suffer from poor generalization due to the distribution gap between the training and the testing distributions.

Finally, subject to the 3D data representation and the availability of labels, several works have proposed different losses to learn their models from \cite{choy20163d, yan2016perspective,kanazawa2018learning,wang2018pixel2mesh, tewari2017mofa,richardson2017learning}. These losses usually present a number of terms related by weighting hyperparameters that need to be tuned for an effective optimization. However, estimating these parameters for each reconstruction dataset is a hard and computationally expensive task that presents high chances of achieving sub-optimal results.

In this work, we propose and study a set of novel losses without hyperparameters for learning model-based monocular reconstruction from real or synthetic data. The main contributions of our work are:
\begin{itemize} 

\item A benchmark of three novel hyperparameter-free losses for learning monocular reconstruction, which have the benefit of decreasing the time and the complexity of the optimization process. We perform an extensive evaluation on an internal large scale 3D dataset and on two public datasets, MICC \cite{bagdanov2011florence} and FaceWarehouse \cite{cao2014facewarehouse}.

\item A novel regularization technique based on random projections that does not require additional 3D or 2D annotations. This allows us to define the Multiview Reprojection Loss (MRL), which is specially suited for those applications that demand a fine-grained alignment between the 3D geometry and the image, such as augmented reality, shape from shading and facial reenactment.

\item An open implementation\footnote{\url{https://github.com/hyperparams-free/hyperparams-free-3D-losses}} of the losses and the 3D annotations used to evaluate the results on MICC \cite{bagdanov2011florence} and FaceWarehouse \cite{cao2014facewarehouse} datasets to facilitate reproducibility and future benchmarkings.

\end{itemize}

The rest of the paper is structured as follows. Section \ref{sec:StateofTheArt} reviews the state of the art for 3D reconstruction from a single image using deep learning models. Section \ref{sec:Hyperparameter-free losses} introduces the three hyperparameter-free losses. Section \ref{sec:Experiments} compares the multiterm losses and the proposed hyperparameter-free losses in terms of performance, robustness and generalization. Finally, Section \ref{sec:Conclusions} draws the conclusions of our work.
\section{State of the art}
\label{sec:StateofTheArt}

Since AlexNet~\cite{krizhevsky2012imagenet} succeeded in training a convolutional neural network (CNN) for large scale image recognition, multiple computer vision tasks have been tackled with deep neural networks~\cite{goodfellow2016deep}. Among them, 3D reconstruction has also benefited from their learned representations, obtaining important performance gains with respect to hand-crafted classic techniques. In general, two big groups of learning-based 3D reconstruction methods can be differentiated by the fact of using or not a 3D morphable model (3DMM), which we will refer as \textit{model-based} and \textit{model-free} approaches respectively.

\subsection{Model-free approaches}
Methods that do not include a 3DMM in their core \cite{yan2016perspective,hane2017hierarchical,wang2018pixel2mesh,jackson2017large,kar2017learning,kanazawa2018learning}, also called \textit{model-free}, are usually oriented to solve generic problems, such as reconstructing objects with different shapes, and are highly conditioned by the 3D representation they use.

For instance, methods based on 3D voxel grids \cite{yan2016perspective,hane2017hierarchical,jackson2017large} tend to use binary cross entropy as objective to optimize their architecture. Eventually, 3D voxel grid geometries can be projected into the image plane to construct supervision signals defined in the image domain, such as depth errors \cite{kar2017learning} or binary masks errors \cite{yan2016perspective}. Despite their flexibility, 3D voxel grid methods are very inefficient at representing surfaces, and hierarchical models are required to achieve denser representations \cite{hane2017hierarchical}. Although they have been mostly assessed in synthetic datasets \cite{shapenet2015}, 3D voxel grid methods have also obtained state of the art results in real applications \cite{jackson2017large}.

Meshes are a common alternative to 3D voxel grids since they are more efficient at surface modelling and have more potential applications. Recent works \cite{kanazawa2018learning,wang2018pixel2mesh} suggest that state of the art results can be achieved by minimizing the Chamfer Loss while regularizing the surface through the Laplace-Beltrami operator and other geometric elements such as normals \cite{wang2018pixel2mesh}. In addition, a family of novel and relevant operators that have been successfully applied to 3D reconstruction with meshes \cite{wang2018pixel2mesh} are the Graph Convolutional Networks (GCN) \cite{bronstein2017geometric}, which generalize the convolution operator to non-Euclidean domains.

\subsection{Model-based approaches}
Model-free methods, specially the mesh based approaches, need to be heavily regularized by using geometric operators in order to obtain plausible 3D reconstructions and, despite their flexibility, are difficult to train. Model-based approaches offer a simpler solution to regularize surfaces by modeling them as a linear combination of a set of basis \cite{blanz1999morphable}. Thus, the learning problem is simplified to estimate a vector of weights to linearly combine the basis of the model.

Due to the lack of 3D data, some works have driven their experiments towards the evaluation of models trained on synthetic data \cite{richardson20163d} \cite{richardson2017learning}. Yet obtaining successful results, iterative error feedback (IEF) \cite{carreira2016human} is usually required for good generalization, which unfortunately implies multiple passes through the network. To speed up the IEF, \cite{kanazawa2017end} performs this process in the latent space. Since using synthetic data provides perfect labels, the losses are designed to explicitly model the error between predictions and ground truth model parameters.

On the other hand, some methods overcome the scarcity of 3D data by defining losses directly in the image domain \cite{tewari2017mofa,tewari2017self,richardson2017learning}. This avoids using IEF since the data is trained and tested in the same distributions. However, annotations on the image domain are required \cite{zhu2016face} or differentiable renderers \cite{kato2017neural} are necessary to construct self-supervised losses using the raw pixel values \cite{tewari2017mofa}. In this case, strong regularization is needed on the predicted model weights to ensure the likelihood of the predicted 3D shapes.

Regularization is a common ingredient in most of the methods used for learning 3D reconstruction. It is usually added as a weighted combination of terms in the loss, either geometric operators for meshes, or norms of the predicted shape model parameters for model-based approaches. These terms provide the model with stability but, at the same time, add complexity to the loss and consequently to the optimization. In \cite{kanazawa2017end}, an adversarial regularization is proposed in order to penalize predicted samples that fall out of the target distribution. This statistical approach is more generic and simpler than using a weighted combination of terms.

Our work follows the direction of \cite{kanazawa2017end} with the objective of finding more generic and simpler losses to learn model-based monocular reconstruction that ease the optimization of the architectures. In contrast to them, we propose different losses based on geometry, instead of statistics, that fuse the data terms and the regularization terms into a single term objective held by the geometry of the problem. As a result, we can dispense with all the hyperparameters.
\section{Hyperparameter-free losses}
\label{sec:Hyperparameter-free losses}

\begin{figure*}
    \centering
    \includegraphics[width=16cm]{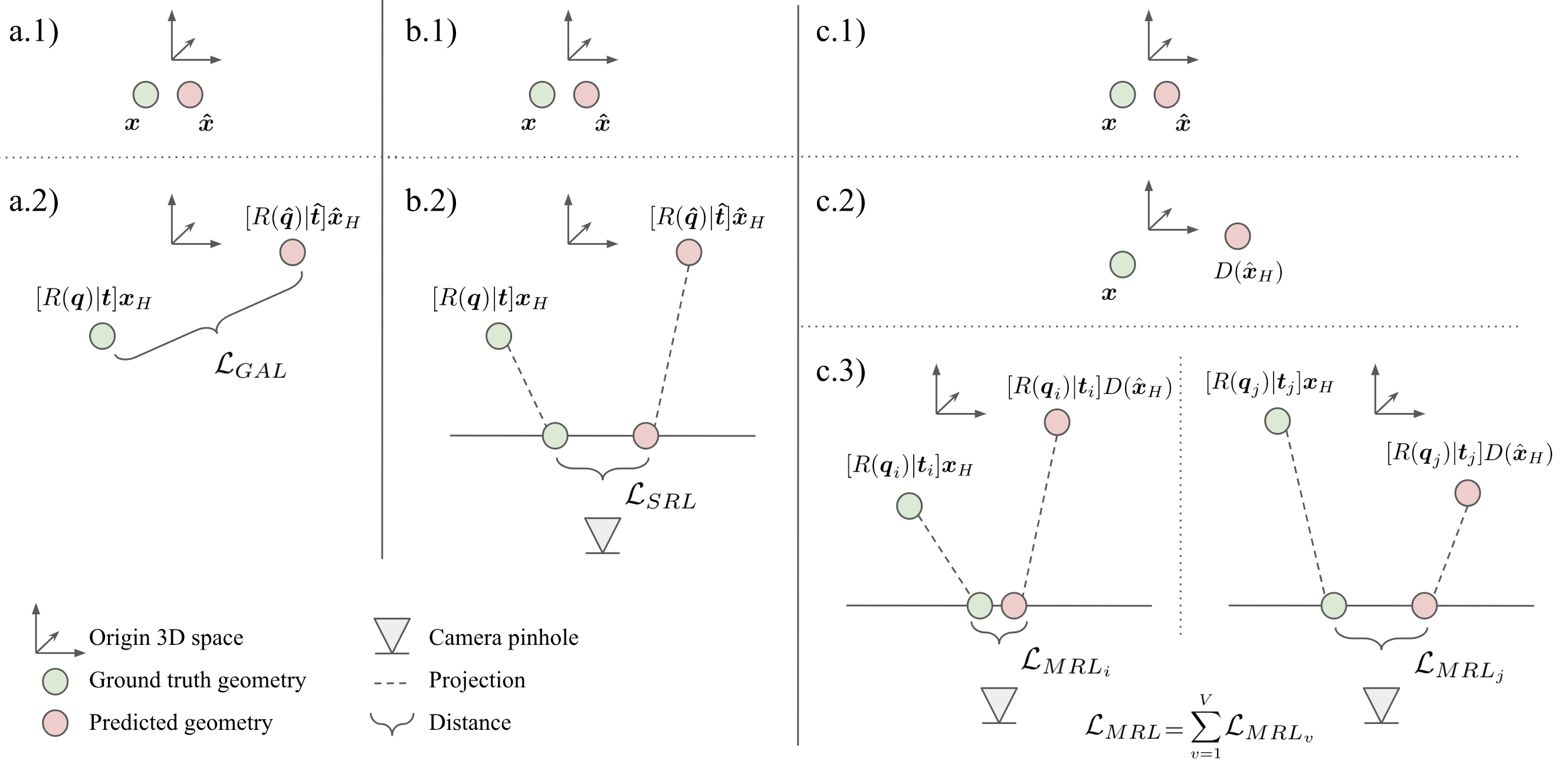}
    \caption{Schemes of the presented hyperparameter-free losses. From top to bottom: Transformations applied to the ground truth and the predictions for computing each loss. From left to right: $\mathcal{L}_{GAL}$ (a), $\mathcal{L}_{SRL}$ (b) and $\mathcal{L}_{MRL}$ (c). The dashed lines represent projections from 3D to the image plane.}
    \label{fig:losses}
\end{figure*}

In this section we introduce three novel hyperparameter-free losses for learning model-based monocular reconstruction. We start by describing the main elements of the problem. Then, we show how the different terms of the losses can be fused into a sole term expression using geometry, which we call Geometric Alignment Loss (GAL). Driven by the fact that a lot of applications require precise alignment between the 3D geometry and the image, we reformulate the GAL loss to minimize the reprojection error, creating the Single View Reprojection Loss (SRL). Finally, we show how the SRL loss can be implicitly regularized through random projections, proposing the last loss called Multiview Reprojection Loss (MRL).

\subsection{Problem statement}\label{subsec:problem_statement}

The problem we address can be defined as finding the unknown mappings from an input image $\mathcal{I}$ to a 3D shape $\boldsymbol{x} \in \mathbb{R} ^{3N} $, N being the number of points, and to the camera pose $c = [ R | \boldsymbol{t} ]$ expressed as a 3x4 matrix, $R$ being the rotation of the camera and $\boldsymbol{t} = ( t_x, t_y, t_z ) \in \mathbb{R}^{3}$ the spatial translation of the camera. We model $R$ as a unit quaternion $\boldsymbol{q} = ( q_0, q_1, q_2, q_3 ) \in \mathbb{H}_1 $ to avoid the Gimbal lock effect, which is the loss of one degree of freedom in a three-dimensional mechanism.

The mappings to be learned can be represented by four functions: $\mathcal{E}$, $\mathcal{X}$, $\mathcal{Q}$ and $\mathcal{T}$. The former function $\mathcal{E}$ is intended to extract relevant features from $\mathcal{I}$ and the rest to map these features to $\boldsymbol{x}$, $\boldsymbol{q}$ and $\boldsymbol{t}$ respectively, so that $\boldsymbol{\hat{x}} = \mathcal{X}( \mathcal{E}( \mathcal{I} ) ) $, $\boldsymbol{\hat{q}} = \mathcal{Q}( \mathcal{E}( \mathcal{I} ) ) $ and $\boldsymbol{\hat{t}} = \mathcal{T}( \mathcal{E}( \mathcal{I} ) ) $ are the predictions of the learnt model.

Most of the current methods based on deep neural networks \cite{tran2017regressing,richardson20163d,richardson2017learning,tewari2017mofa,tewari2017self} learn the mapping functions $\mathcal{E}$, $\mathcal{X}$, $\mathcal{Q}$ and $\mathcal{T}$ by linearly combining different loss terms. Each of these terms is responsible for controlling a property of the reconstruction, and its contribution to the final loss is adjusted by a weighting hyperparameter that must be tuned. 
In general, these loss terms can be divided in data terms and regularization terms \cite{tewari2017mofa}. 

Data terms are the ones that guide the network predictions, $\boldsymbol{\hat{x}}$, $\boldsymbol{\hat{q}}$ and $\boldsymbol{\hat{t}}$, towards matching the ground truth labels $\boldsymbol{x}$, $\boldsymbol{q}$ and $\boldsymbol{t}$ during training:
\begin{equation}\label{eq:multiterm_data}
\mathcal{L}_{data} = \mathcal{L}_{\hat{x}} + \alpha\mathcal{L}_{\hat{q}} + \beta \mathcal{L}_{\hat{t}}.
\end{equation}

As noted in \cite{kendall2017geometric}, the relation between the hyperparameters $\alpha$ and  $\beta$ varies substantially depending on the problem and, consequently, the choice of these hyperparameters has a severe impact for the camera pose estimation.

On the other hand, regularization controls the predicted 3D shape $\boldsymbol{\hat{x}}$ in terms of geometric and semantic likelihoods. In this sense, it is common to use a 3DMM, which allows to represent the predicted geometry in a lower dimensional space. More precisely, it expresses $\boldsymbol{\hat{x}}$ as:

\begin{equation}\label{eq:3DMM}
 \boldsymbol{\hat{x}} =  \boldsymbol{m} + \Phi_{id} \boldsymbol{\hat{\alpha}}_{id},
\end{equation}
where $\boldsymbol{m}$ represents the mean of the 3DMM, and  $\Phi_{id}$ and $\boldsymbol{\hat{\alpha}}_{id}$ are the identity basis and the predicted identity parameters respectively.

In order to obtain plausible shapes, $\boldsymbol{\hat{\alpha}}_{id}$ needs to have a small norm. Consequently, those methods that do not have access to 3D ground truth or that define their losses entirely in the image domain \cite{tewari2017mofa} must include an extra regularization term in their loss that force this condition during training:

\begin{equation}
\mathcal{L}_{reg} = \gamma || \boldsymbol {\hat{\alpha}}_{id} ||_2^2.
\end{equation}

A typical hyperparameter-dependent loss would simply sum the data and regularization terms:

\begin{equation}
\mathcal{L}=  \mathcal{L}_{data}  + \mathcal{L}_{reg}.
\end{equation}

In general, methods that learn monocular reconstruction define their losses following the described multiterm strategy, which require an estimate of the weighting hyperparameters $\alpha$ and $\beta$ for each specific dataset, a hard and expensive process that might lead to suboptimal results.

From now on, we assume that the 3D shape can be expressed using a 3DMM as in Equation \ref{eq:3DMM}, and that real or synthetic 3D ground truth is available.

\subsection{Using geometry to avoid the hyperparameters}

In this section, we propose a simple but effective reformulation of the standard multiterm losses (Equation \ref{eq:multiterm_data}) that unifies the errors produced by $\boldsymbol{\hat{x}}$, $\boldsymbol{\hat{q}}$ and $\boldsymbol{\hat{t}}$ into a single term expression. We call this formulation \textit{Geometric Alignment Loss} (GAL) and it is defined as follows:

\begin{align}\label{eq:geometric_alignment_loss}
\mathcal{L}_{GAL} =  || [ R(\boldsymbol{ q }) | \boldsymbol{ t } ] \boldsymbol{ x }_H - [ R(\boldsymbol{ \hat{q} }) | \boldsymbol{ \hat{t} } ] \boldsymbol{ \hat{x} }_H ||_1,
\end{align}
$R(\boldsymbol{ q })$ being the rotation matrix induced by the quaternion $\boldsymbol{ q }$, and $\boldsymbol{ x }_H$ the 3D shape in homogeneous coordinates.

Essentially, $\mathcal{L}_{GAL}$ uses the rotation and the translation of the camera pose to align the ground truth shape and the predicted shape in the 3D space, and then compute point to point distances. This process is illustrated in Figure \ref{fig:losses} a). From our experiments, we find $\ell_1$ norm to behave the best in terms of stability and accuracy. Note that the surface of the loss is well defined, since the use of a 3DMM constrains the position and the orientation of the predicted 3D shape, avoiding possible ambiguities in the product between $[ R(\boldsymbol{ \hat{q} }) | \boldsymbol{ \hat{t} } ]$ and $ \boldsymbol{ \hat{x} }_H$.

\subsection{Reprojection error as objective}

Obtaining an accurate shape and camera pose is, by definition, the goal of single view 3D reconstruction. However, a number of applications such as texture generation, face reenactment, augmented reality and shape from shading based geometry refinement, specially demand a precise alignment between the predicted geometry $\boldsymbol{ \hat{x} }$ and the input image $\mathcal{I}$. Although it might result unintuitive, small errors in the camera rotation and the camera translation, do not necessarily imply low reprojection errors, since they can compensate or aggregate each other.

Despite GAL already avoids the use of hyperparameters, we would like to obtain a unique term formulation that not only optimizes shape and pose simultaneously, but that it also achieves the lowest possible reprojection error for those applications that require fine-grained alignment between 2D and 3D spaces.

We get inspiration from \cite{kendall2017geometric}, where the camera pose is estimated by minimizing the reprojection error, and we introduce the predicted geometry to define the \textit{Single View Reprojection Loss} (SRL), which is illustrated in Figure \ref{fig:losses} b):

\begin{equation}\label{eq:single_view_reprojection_error}
\mathcal{L}_{SRL} = || \mathcal{P}( \boldsymbol{ q }, \boldsymbol{ t } )( \boldsymbol{ x }_H ) - \mathcal{P}(  \boldsymbol{ \hat{q}}, \boldsymbol{ \hat{t} } )( \hat{\boldsymbol{x}}_H ) ||_1,
\end{equation}
where $\mathcal{P}$ projects any 3D shape $\boldsymbol{y}$ to the 2D image plane, obtaining $\boldsymbol{y}_{2D}$ defined by:

\begin{equation}
\boldsymbol{ y }_{2D} = \begin{pmatrix} u' / w' \\ v' / w' \end{pmatrix},
\end{equation}
with

\begin{equation}\label{eq:projective transform}
\begin{pmatrix} u' v' w' \end{pmatrix}^{T} = K [ R(\boldsymbol{ q }) | \boldsymbol{ t } ] \boldsymbol{ y }_H,
\end{equation}
$K$ being the calibration matrix.

\begin{figure}
    \centering
    \includegraphics[width=8cm]{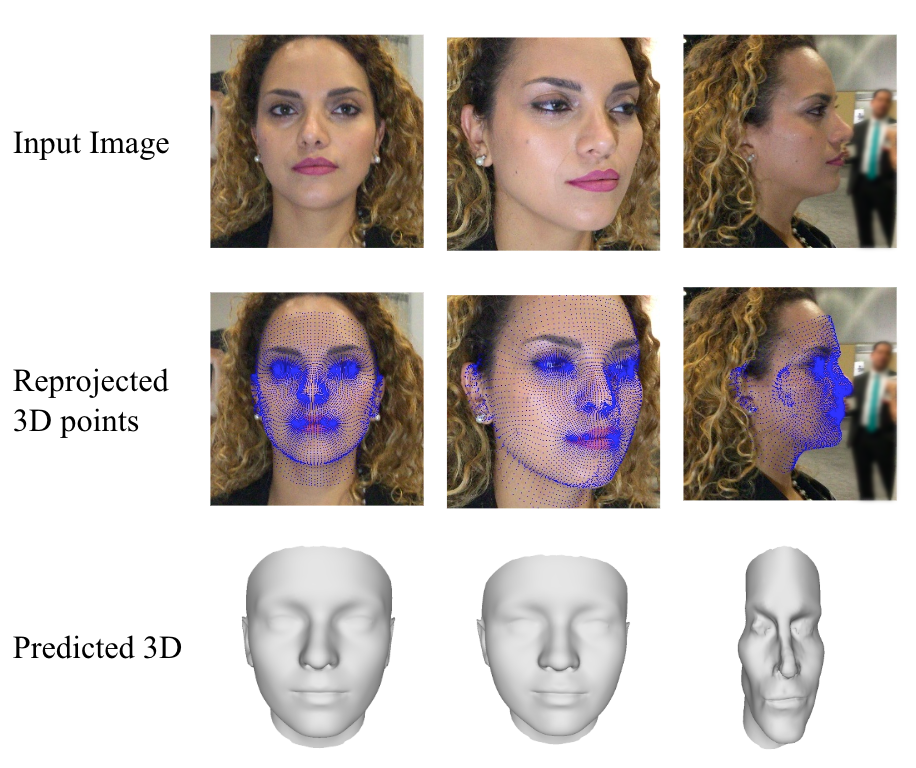}
    \caption{Effect of training with SRL. While the reprojection error is minimized, the 3D shape is not plausible. }
    \label{fig:flattening}
\end{figure}

By using the SRL loss, one can simultaneously learn shape and pose by minimizing the reprojection error. Unfortunately, as commented in Section \ref{subsec:problem_statement}, optimizing 3D shape and pose by projecting into a single image plane is not possible without regularization. As it can be observed in Figure \ref{fig:flattening}, the network learns to generate flattened shapes $\hat{\boldsymbol{x}}$ in the profile views, which produce minimum reprojection error but do not belong to the distribution of geometrically plausible 3D faces.

\subsection{Implicit regularization via random projections}\label{subsec:impl_reg}

A trivial solution to regularize the predictions of $\hat{\boldsymbol{x}}$ and avoid the flattened shapes produced by SRL would be to add an extra term, $||\boldsymbol{\hat{\alpha}}_{id}||_2^2$, to Equation \ref{eq:single_view_reprojection_error} in order to keep the norm of $\boldsymbol{\hat{\alpha}}_{id}$ small. This would introduce an extra hyperparameter that we would like to avoid. 

Instead, we propose to implicitly regularize the learning process of $\boldsymbol{ \hat{x} }$  by projecting it to multiple random image planes. The error produced by $\boldsymbol{ \hat{q} }$ and $\boldsymbol{ \hat{t} }$ is introduced as an isometric transform $D$ that distorts the predicted geometry $\boldsymbol{ \hat{x} }$ in position and orientation. Then, we define the \textit{Multiview Reprojection Loss} (MRL) as:

\begin{equation}\label{eq:mrl}
\mathcal{L}_{MRL} = \sum_{v=1}^{V} || \mathcal{P}( \boldsymbol{ q_v }, \boldsymbol{ t_v } )( \boldsymbol{ x }_H ) - \mathcal{P}(  \boldsymbol{ q_v}, \boldsymbol{ t_v } )( D(\hat{\boldsymbol{x}}_H) ) ||_1,
\end{equation}
where $\boldsymbol{ q_v }$ and $\boldsymbol{ t_v }$ represent the camera pose of a random view. The isometric transform $D$ is defined as the relative pose between the predicted camera pose and the ground truth camera pose expressed as 4x4 matrices:

\begin{equation}\label{eq:distortion}
D(\hat{\boldsymbol{x}}_H) = [ R(\boldsymbol{ q }) | \boldsymbol{ t } ]\cdot [ R(\boldsymbol{ \hat{q} }) | \boldsymbol{ \hat{t} } ]^{-1} \hat{\boldsymbol{x}}_H.
\end{equation}

The MRL allows to simultaneously learn the 3D shape and the camera pose without explicit regularization of $\hat{\boldsymbol{x}}$ and, at the same time, achieves minimum reprojection errors. We illustrate it in Figure \ref{fig:losses} c).
\section{Experiments}
\label{sec:Experiments}

We evaluate the losses presented in Section \ref{sec:Hyperparameter-free losses} in terms of accuracy, robustness, efficiency and generalization. In order to isolate at maximum the effects of each loss, we use the same architecture and the same training data to optimize all the models, as well as the same testing data for evaluation. The only difference between configurations is the loss function used during training.

\subsection{Dataset}
\label{ssec:Dataset}

One of the main challenges for learning 3D reconstruction models is the scarcity of 3D annotations. Strategies to overcome this issue range from using synthetic data \cite{richardson20163d,richardson2017learning,guo2018cnn} to fitting 3DMM to images \cite{zhu2016face,feng2018joint}. However, the 3D ground truth produced by these strategies is subject to inaccuracies in the input data distribution caused by the renderers or in the target geometry caused by the fittings of the 3DMM. To the best of our knowledge, there are not publicly available datasets with real images and accurate 3D ground truth large enough for the training and evaluation of single view 3D reconstruction models.

\begin{figure}
    \centering
    \includegraphics[width=8cm]{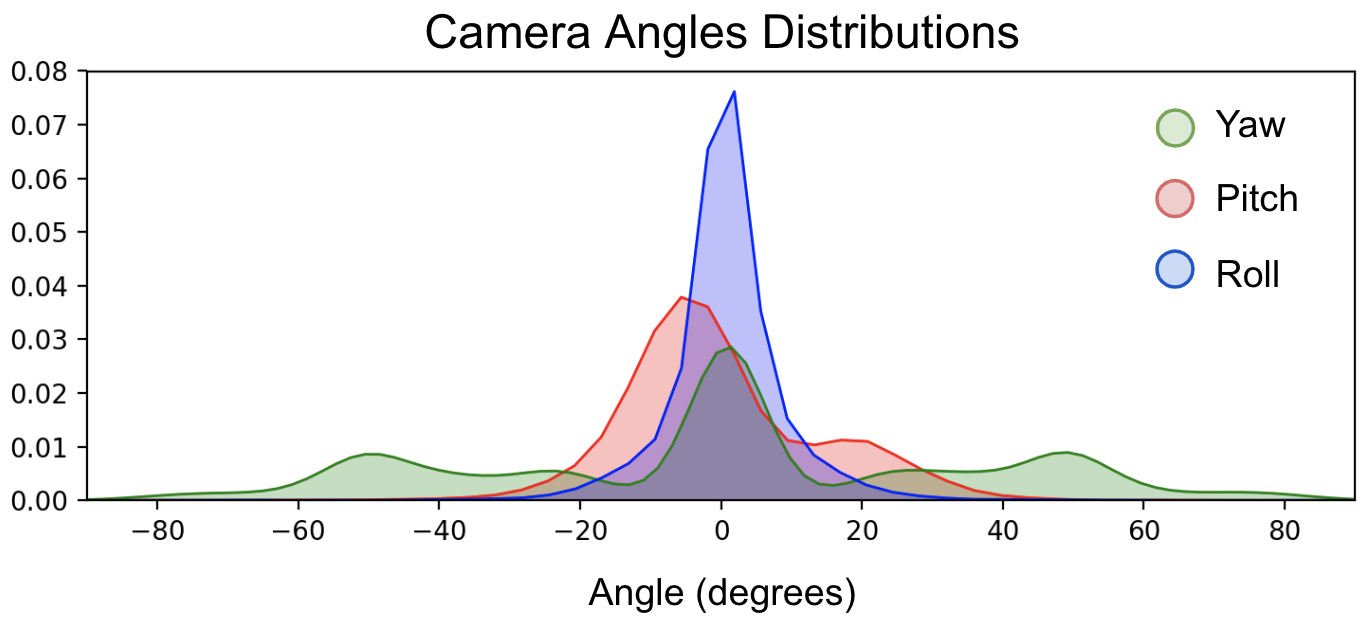}
    \caption{Camera angles distributions.}
    \label{fig:yaw_pitch_roll}
\end{figure}

In order to be as rigorous as possible, we built a large scale 3D dataset with real images and accurate 3D ground truth. Concretely, we scan a total of 6528 individuals from different gender, age and ethnicity. From each subject, we acquire the facial geometry without expressions using the Structure Sensor scanner from Occipital. We also obtain multiple RGB images and their respective camera poses from multiple views. All the scenes are normalized so that the heads are aligned towards a reference 3D template, which is centered at $\boldsymbol{\vec{0}}$ and facing towards -$\boldsymbol{\hat{z}}$. We separate the subjects in three subgroups, train, validation and test, using approximately the 70\%, 10\% and 20\% of the data respectively.  Table \ref{tab:dataset} shows the numerical details of the data partitions used for training, validation and testing, and Figure \ref{fig:yaw_pitch_roll} the camera angle distributions. For data augmentation purposes, each scan and its respective images and camera poses are fully symmetrized.

\begin{table}[]
\centering
\begin{tabular}{l|c|c|c}
\toprule
& & &  Average \\
Split  & \# subjects & \# images & views/subject \\
\midrule
Train &  4543 & 20349 & 4.4 \\
Validation &  675 & 2976 & 4.4 \\
Test &  1310 & 6347 & 4.8 \\
\bottomrule
\end{tabular}
\caption{Dataset details for training, validation and testing.}
\label{tab:dataset}
\end{table}

Finally, in order to create the 3DMM, we register the 3D reference template to the 3D scans from the training set using a Non-Rigid ICP algorithm. Then, Procrustes analysis is performed using all the registered models, and Principal Component Analysis (PCA) is applied to extract the identity bases $\Phi_{id}$ and the associated eigenvalues $\Lambda$.

This dataset provides us with enough data to train and evaluate deep architectures with the necessary precision to extract solid conclusions from our experiments.

\subsection{Implementation details}
\label{ssec:Implementation}

We select a standard architecture to predict the first 100 identity parameters $\boldsymbol{\hat{\alpha}}_{id}$ of the 3DMM, the camera rotation as a unit quaternion $\boldsymbol{\hat{q}} = ( \hat{q}_0, \hat{q}_1, \hat{q}_2, \hat{q}_3 )$, and the spatial camera translation $\boldsymbol{\hat{t}} = ( \hat{t}_x, \hat{t}_y, \hat{t}_z )$. Similarly to \cite{richardson20163d,tewari2017mofa,tewari2017self,richardson2017learning} we choose a convolutional neural network as encoder $\mathcal{E}$ based on VGG-16 \cite{simonyan2014very} to extract image features, and three multilayer perceptrons (MLP), $\mathcal{S}$, $\mathcal{Q}$ and $\mathcal{T}$, with 1 hidden layer of 256 units, that are added on top of $\mathcal{E}$ to regress $\boldsymbol{\hat{\alpha}}_{id}$, $\boldsymbol{\hat{q}}$ and $\boldsymbol{\hat{t}}$ respectively. Since the set of 3D rotations is represented by quaternions of norm 1, we add a normalization layer to the quaternion branch, being the final mapping $\mathcal{\bar{Q}} = \mathcal{Q} / ||\mathcal{Q}||_2$. Moreover, we add a frozen linear layer on top of $\mathcal{S}$ to directly predict the 3D geometry $\boldsymbol{\hat{x}}$ from  $\boldsymbol{\hat{\alpha}}_{id}$ as shown in Equation \ref{eq:3DMM}, obtaining the final mapping $\mathcal{X} = \boldsymbol{m} + \Phi_{id} \mathcal{S}$.

Given an input image $\mathcal{I}$, the three outputs of our model can be expressed as: $\boldsymbol{\hat{x}} = \mathcal{X}( \mathcal{S}(\mathcal{E}( \mathcal{I} ) ) ) $, $\boldsymbol{\hat{q}} = \overline{\mathcal{Q}( \mathcal{E}( \mathcal{I} ) )} $ and $\boldsymbol{\hat{t}} = \mathcal{T}( \mathcal{E}( \mathcal{I} ) ) $. For better initial conditions, we initialize the layers $\mathcal{S}$, $\mathcal{Q}$ and $\mathcal{T}$ in order to predict $\boldsymbol{\hat{\alpha}}_{id}=\vec{0}$, $\hat{\boldsymbol{q}}=[1,0,0,0]$ and $\hat{\boldsymbol{t}}=[ 0, 0, -60 ]$, values that project the mean 3D shape to the center of the image. Unless differently specified, all the models have been trained until convergence using Adam \cite{kingma2014adam} with a learning rate of $10^{-4}$ and batch size of 32 samples on a NVIDIA RTX 2080 Ti.

\subsection{Metrics}
\label{ssec:metrics}

We use different metrics to quantify the prediction errors of the 3D shape, the camera translation, the camera rotation and the reprojected shapes. Here, we rapidly formalize how these errors are computed for each subject as well as the units:

\begin{itemize}
    \item Shape 3D error (mm): $\sum_{n=1}^{N_p} || \boldsymbol{x}_n - \boldsymbol{\hat{x}}_n ||_2 / N_p$
    \item Camera translation error (cm): $|| \boldsymbol{t} - \boldsymbol{\hat{t}} ||_2$
    \item Camera rotation error (degrees): $acos( 2 \boldsymbol{q} \cdot \boldsymbol{\hat{q}})·180 / \pi $
    \item Reprojection error (pixels): \\
    $\sum_{n=1}^{N_p} || \mathcal{P}( \boldsymbol{ q }, \boldsymbol{ t } )( {\boldsymbol{x}_{n}}_{H}) - \mathcal{P}( \boldsymbol{ \hat{q} }, \boldsymbol{ \hat{t} } )( {\boldsymbol{\hat{x}}_{n}}_{H} ) ||_2 / N_p$,
\end{itemize}
where $N_p$ is the number of points in the 3D shape and $x_{n} \in \mathbb{R} ^{3}$ is the nth point of the 3D shape.

\subsection{Quantitative evaluation}
\label{ssec:quantitative}

In this section we compare the performance of the multiterm losses against the hyperparameter-free ones.
To begin with, we implement the multiterm loss described in the state of the art work \cite{richardson2017learning}, since it also uses 3D annotations but synthetically generated: 
\vspace{-1em}

\begin{equation}\label{eq:coarse}
\mathcal{L}_{Coarse} = || \boldsymbol{x} - \boldsymbol{\hat{x}} ||^2_2 + \alpha||  [ \boldsymbol{q}, \boldsymbol{t} ] - [ \hat{ \boldsymbol{q}}, \hat{ \boldsymbol{t}} ] ||^2_2,
\end{equation}
where $[ \cdot , \cdot ]$ is the concatenation operator. Note that the only difference with respect to \cite{richardson2017learning} is that we are assuming a pinhole camera model instead of a weak perspective model.

The $\mathcal{L}_{Coarse}$ does not balance the errors produced by $\hat{ \boldsymbol{q}}$ and $\hat{ \boldsymbol{t}}$. For completeness, as \cite{kendall2017geometric} shows the importance of having two weighted terms for $\boldsymbol{\hat{q}}$ and $\boldsymbol{\hat{t}}$, we also implement and evaluate the following multiterm expression:

\vspace{-1em}

\begin{equation}\label{eq:xqt}
\mathcal{L}_{XQT} = || \boldsymbol{x} - \boldsymbol{\hat{x}} ||^2_2 + \beta|| \boldsymbol{q} - \hat{ \boldsymbol{q}} ||^2_2 + \gamma||  \boldsymbol{t} - \hat{ \boldsymbol{t}} ||^2_2,
\end{equation}
which can be understood as the combination of the Geometric Mean Squared Error (GMSE) defined in \cite{richardson20163d} and used for learning the geometry, and the cost defined in \cite{kendall2015posenet} and used for learning the camera pose.

The best models trained with $\mathcal{L}_{Coarse}$ and $\mathcal{L}_{XQT}$ are obtained after a Bayesian optimization to estimate the learning rate and ${\alpha}$ and $\{\beta, \gamma\}$, respectively. To find the search space bounds, we estimate the $\alpha$, $\beta$ and $\gamma$ values that compensate the difference of scale with the term $|| \boldsymbol{x} - \boldsymbol{\hat{x}} ||^2_2$ as in \cite{richardson2017learning}, obtaining $\alpha_{scale}$, $\beta_{scale}$ and $\gamma_{scale}$. Then, the lower and the upper bounds of the search space are defined by an order of magnitude below and an order of magnitude above the estimated values: $\alpha_{opt} \in ( 0.1 \alpha_{scale}, 10 \alpha_{scale})$, $\beta_{opt} \in ( 0.1 \beta_{scale}, 10 \beta_{scale})$ and $\gamma_{opt} \in ( 0.1 \gamma_{scale}, 10 \gamma_{scale})$. Regarding the learning rate, we define the search interval as $(10^{-5}, 10^{-3})$. We also limit the Bayesian optimization search to 20 experiments.

On the other hand, we train three more models using the proposed hyperparameter-free losses, $\mathcal{L}_{GAL}$, $\mathcal{L}_{SRL}$ and $\mathcal{L}_{MRL}$, with the learning rate fixed to $10^{-4}$. In this case, the training is performed a single time.

\begin{table*}[]
\centering
\begin{tabular}{@{}lcccccccc@{}}
\toprule
    & Repro- & Shape  & Camera  &  Camera &  &  &  & \\
Loss& jection& 3D  & translation  &  rotation &  Time/epoch & Epochs & Trainings & Total time \\
    & (pixels) & (mm)  & (cm)  &  (degrees) & (minutes) &  & & (days) \\
\midrule
Coarse \cite{richardson2017learning} & 11.1 & \textbf{2.3} & \textbf{3.0}  & \textbf{3.0} & \textbf{6.8} & 120 & 20 & 11.3 \\
XQT & 11.6 &  2.5 &  3.3 &  3.1 & 6.9 & 120 & 20 & 11.5 \\
\midrule
GAL & 17.1 &  2.8 &  \textbf{3.0} &  3.1 & 9.2 & 120 & 1 & \textbf{0.8} \\
SRL & \textbf{3.3} & 9.0  & 12.6 &  51.7 & 9.3 & 500 & 1 & 3.2 \\
MRL (2 views) & 4.3 & 3.0 & 4.2 & 4.3 & 14.9 & 120 & 1 & 1.2 \\
\bottomrule
\end{tabular}
\caption{
Performance comparison of the models trained with the different losses. Top: multiterm losses with optimal parameters found using Bayesian optimization. Bottom: Hyperparameter-free losses trained a single time.
 }
\label{tab:quantitative_results}
\end{table*}

Table \ref{tab:quantitative_results} shows the quantitative results obtained after training the models and evaluating them on our dataset. As it can be observed, hyperparameter-free losses allow a much faster optimization process while obtaining comparable accuracies. Moreover, the SRL and the MRL obtain much lower reprojection errors than the optimized multiterm losses, but only MRL is capable to achieve a good balance between the reprojection error and the 3D shape error due to the implicit regularization. On the other hand, the optimized multiterm models obtain slightly better results (tenths of a millimeter) in terms of 3D shape accuracy and in terms of camera pose estimation with respect GAL and MRL.

\subsection{Robustness against large poses}
\label{ssec:robustness}

\begin{figure}
    \centering
    \includegraphics[width=8cm]{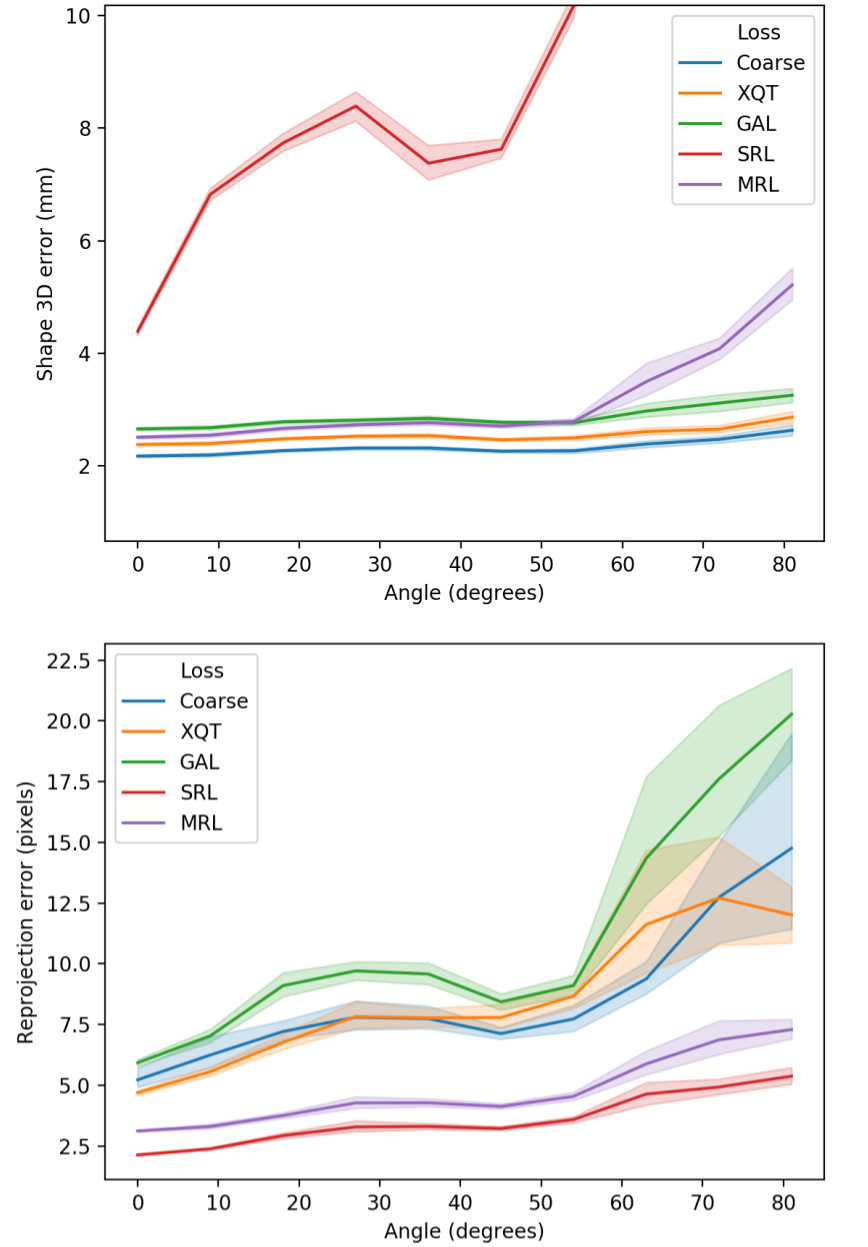}
    \caption{Shape 3D errors (top) and reprojection errors (bottom) depending on camera angles.}
    \label{fig:per_angle}
\end{figure}

It is also interesting to observe how the models trained with the different losses behave depending on the camera angle, which we plot in Figure \ref{fig:per_angle}. This fact is tightly related with the abundance of data shown in Figure \ref{fig:yaw_pitch_roll}. The multiterm losses and GAL generalize better than SRL and MRL to predict the 3D shape for large posses, where the information is poorer, but fail to achieve stable reprojection errors. On the opposite side, SRL and MRL provide much more robust predictions in terms of reprojection error, but only MRL achieves a reasonable stability in terms of 3D shape.

\subsection{Random projections in MRL}
\label{ssec:n_in_mrl}

\begin{figure}
    \centering
    \includegraphics[width=8cm]{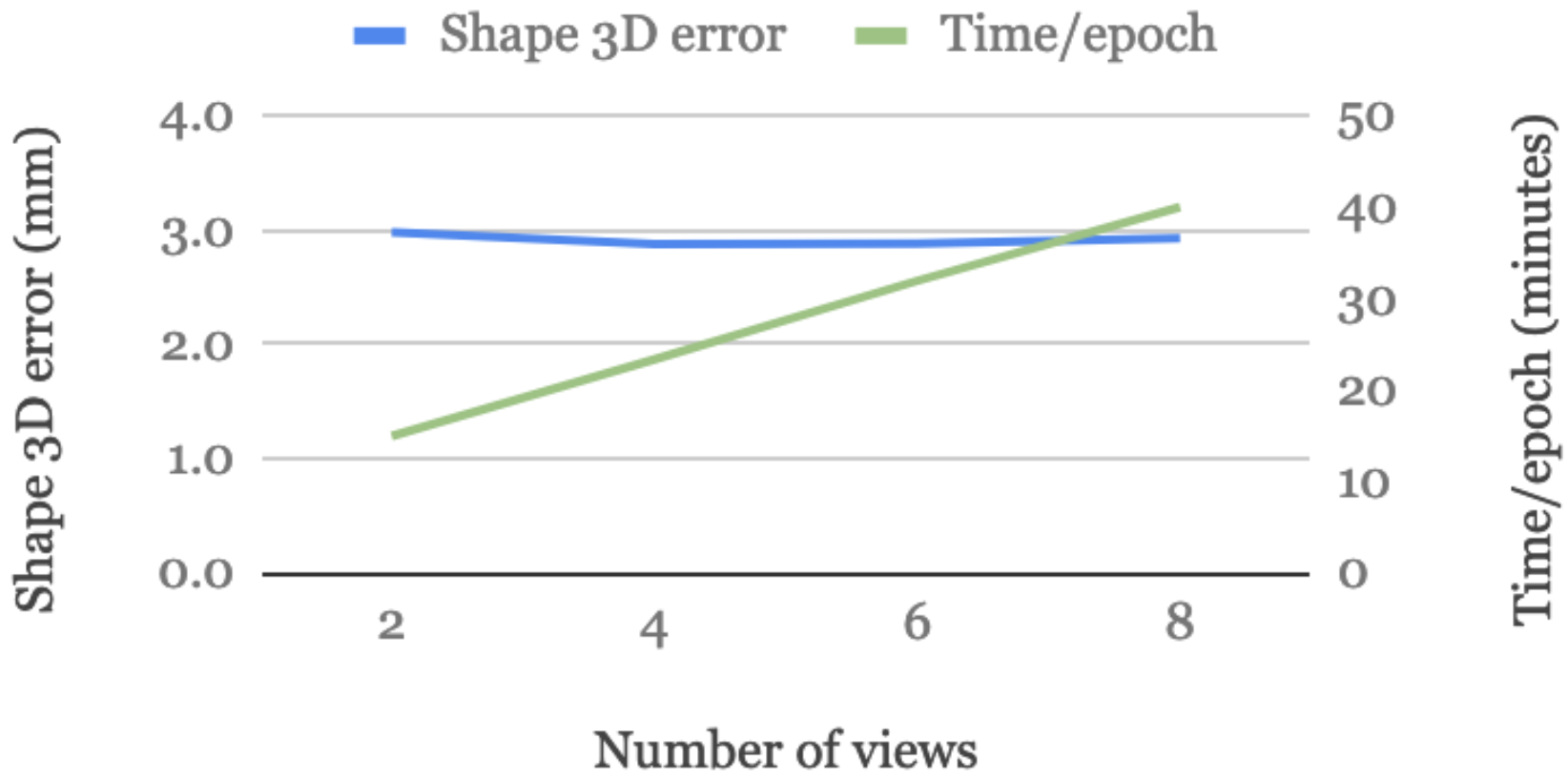}
    \caption{Effect of the number of views on the models trained using MRL.}
    \label{fig:performance_vs_time}
\end{figure}

Using multiple random views allows the MRL to regularize the predictions of the 3D shape. Figure \ref{fig:performance_vs_time} shows that the variations in the shape 3D error are smaller than a tenth of millimeter and therefore can be considered negligible. On the other hand, the computational cost grows linearly with the number of views. From these results, we conclude that using $V=2$ is sufficient to train accurate and stable models.

\subsection{Generalization to other datasets.}
\label{ssec:generalization}

In order to measure how each loss contributes to the generalization, we evaluate the five models from Section \ref{ssec:quantitative} on the MICC \cite{bagdanov2011florence} and the FaceWarehouse \cite{cao2014facewarehouse} datasets. Since our training set only contains faces with neutral expressions, we perform inference on the subset of images from each MICC and FaceWarehouse without expressions. Moreover, on MICC we select the most frontal frame for each subject in order to match the ground truth geometry as much as possible. Once the 3D shape is predicted, it is aligned towards the 3D ground truth using manually annotated 3D landmarks and performing Iterative Closest Point (ICP), as in \cite{tran2017regressing}. We publish the selected frames and the manually annotated 3D landmarks in the provided repository to allow reproducibility.

\begin{figure}
    \centering
    \includegraphics[width=8cm]{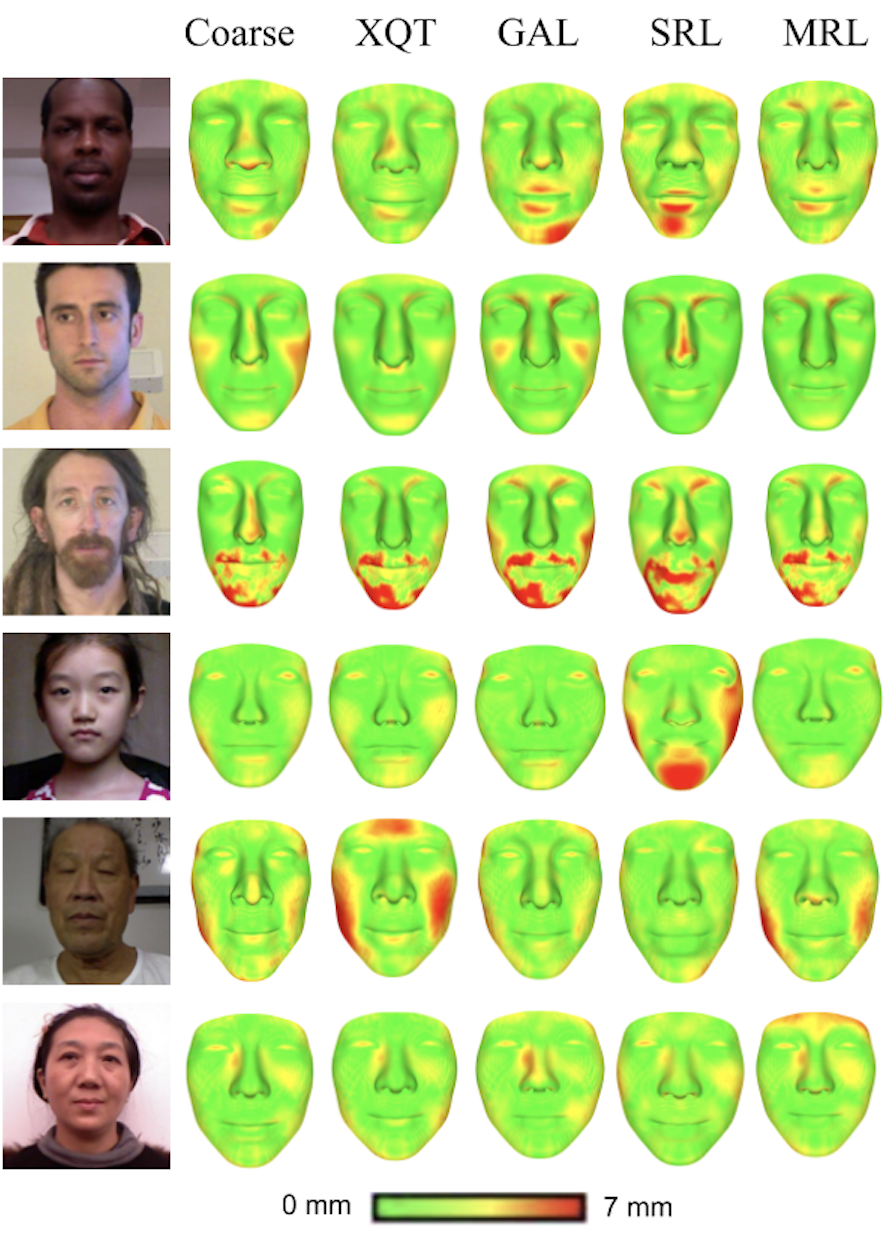}
    \caption{Qualitative evaluation of the shape 3D errors on cases from MICC and FaceWarehouse.}
    \label{fig:heatmaps}
\end{figure}

As it can be observed in Table \ref{tab:generalization}, multiterm losses obtain similar shape 3D errors to the ones reported in Table \ref{tab:quantitative_results} and  Figure \ref{fig:per_angle}. However, the gap in performance between the multiterm losses and the hyperparameter-free losses has been reduced in MICC and FaceWarehouse, specially for GAL and MRL. This suggests that GAL and MRL generalize better to unseen data distributions than the multiterm losses. Figure \ref{fig:heatmaps} provides qualitative evidences that the shape 3D errors are similar, specially within the models trained with Coarse, XQT, GAL and MRL losses.

\begin{table}[]
\centering
\begin{tabular}{@{}lcc@{}}
\toprule
    & MICC \cite{bagdanov2011florence} & FaceWarehouse \cite{cao2014facewarehouse} \\
\midrule
Coarse \cite{richardson2017learning} & \textbf{2.2} & \textbf{2.2}  \\
XQT & 2.3 &  \textbf{2.2} \\
\midrule
GAL & \textbf{2.2} &  \textbf{2.2} \\
SRL & 2.9 &  2.8 \\
MRL & \textbf{2.2} & 2.3  \\
\bottomrule
\end{tabular}
\caption{ Shape 3D error in millimeters computed on MICC and FaceWarehouse datasets.}
\label{tab:generalization}
\end{table}

\section{Conclusions}\label{sec:Conclusions}

We have introduced three novel hyperparameter-free losses for model-based monocular reconstruction. Our experiments suggest that, by using these losses instead of the multiterm ones, the complexity and the time spent on optimizing the models is considerably reduced while achieving comparable accuracy, robustness and generalization.

The SRL performs the best at minimizing the reprojection error but the lack of regularization produces unstable 3D shape predictions, specially for large poses. The GAL loss is more stable in terms of shape 3D error against large posses, similarly to the multiterm approaches, and it allows to rapidly obtain competitive models. In contrast, the MRL is a bit more slow than GAL but it shows much more stability in the reprojection error, making it suitable for applications that require fine-grained alignment between image and geometry such as augmented reality.

Considering these advantages, we conclude that both GAL and MRL are great alternatives to the multiterm losses for learning model-based monocular reconstruction.

{\small
\bibliographystyle{ieee}
\bibliography{paper}
}

\end{document}